\documentclass[runningheads]{llncs}
\usepackage[T1]{fontenc}
\usepackage{graphicx,verbatim}
\usepackage{amsmath}
\usepackage{amssymb}
\usepackage{booktabs}
\usepackage{array}
\usepackage{multirow}
\usepackage{microtype}

\begin{document}

\title{Anatomy-Slot: Unsupervised Anatomical Factorization for Homologous Bilateral Reasoning in Retinal Diagnosis}
\titlerunning{Anatomy-Slot for Bilateral Retinal Diagnosis}


\author{Yingzhe Ma\inst{1}$^{\star}$ \and
Xiao Yang\inst{1}$^{\star}$ \and
Yuguo Yin\inst{2} \and
Zheyu Wang\inst{1}}
\authorrunning{Y. Ma et al.}
\institute{University of Electronic Science and Technology of China \and
Peking University\\
\noindent$^{\star}$ Equal contribution.}

\maketitle

\begin{abstract}
Retinal diagnosis is inherently bilateral: clinicians compare homologous structures across eyes (e.g., optic disc asymmetry), yet most deep models operate on monocular representations. We investigate whether explicit structural correspondence improves diagnosis, and propose \textbf{Anatomy-Slot} to operationalize this hypothesis. Anatomy-Slot introduces an unsupervised anatomical bottleneck by decomposing patch tokens into a set of emergent, structurally-coherent slots that correspond to anatomical regions, then aligning these slots across eyes via bidirectional cross-attention. On ODIR-5K with \(n=10\) seeds, the method improves AUC by $\mathbf{4.2}$ points over a matched ViT-L baseline (95\% CIs; Wilcoxon signed-rank test, $W=0$, $p=0.002$). Pairing disruption and stress testing under Gaussian noise provide controlled tests of correspondence dependence and robustness under corruption. We further report quantitative optic disc grounding on REFUGE and cross-attention localization analysis. Beyond the reported gains, these results indicate that object-centric anatomical correspondence offers a principled path toward interpretable diagnostic systems aligned with clinical bilateral comparison.

\end{abstract}

\section{Introduction}
Retinal imaging is a key screening tool for diabetic retinopathy, glaucoma, and age-related macular degeneration. In clinical practice, retinal assessment is fundamentally bilateral: clinicians compare homologous structures---optic disc, macula, vasculature, peripheral retina---across eyes, where asymmetries in disc cupping, vessel caliber, or lesion distribution provide early diagnostic signals. Despite this, most deep models process each eye independently, discarding the bilateral context clinicians rely upon. We study multi-label diagnosis over eight categories (Normal, Diabetes, Glaucoma, Cataract, AMD, Hypertension, Myopia, Other), where bilateral comparison is clinically routine yet variably informative across conditions. Throughout, ``unsupervised'' refers to slot decomposition and correspondence learning; the overall model uses image-level labels for classification.

With over 2.2 billion people affected by vision impairment globally~\cite{ref_who_vision} and a growing shortage of specialist readers, automated screening aligned with clinical workflow is urgently needed. Yet a fundamental question remains: can models explicitly reason about cross-eye anatomical correspondence, and does this yield measurable diagnostic gains? Prior bilateral architectures via Siamese networks or dual-branch fusion~\cite{ref_bilateral1,ref_bilateral2,ref_bilateral3,ref_oct1} aggregate paired information through late fusion or symmetric constraints without representing \emph{which} structures are compared across eyes. We argue that structured correspondence---mapping optic disc to optic disc, macula to macula, vessel arcades to their contralateral counterparts---is the key missing ingredient.

We propose \textbf{Anatomy-Slot}, which decomposes each fundus image into object-centric slots via Slot Attention~\cite{ref_slot} and aligns slots across eyes through bidirectional cross-attention. Slots serve as unsupervised correspondence units; cross-attention enables each eye to query homologous structures in the contralateral eye. This combination is non-trivial: direct cross-attention over patch tokens yields diffuse, unstructured patterns, whereas the slot bottleneck forces specialization to coherent anatomical zones, enabling cross-attention over semantically meaningful units. We investigate two research questions:

\noindent\textbf{(RQ1)} Does explicit cross-eye correspondence modeling with object-centric units improve diagnostic performance under matched-backbone, statistically controlled settings?

\noindent\textbf{(RQ2)} Do such models exhibit behavioral signatures consistent with correspondence reliance---specifically, degradation when pairing is disrupted?

We evaluate with \(n=10\) independent seeds, 95\% CIs, and non-parametric paired testing, complemented by mechanistic analyses (pairing disruption, stress testing, Fisher Ratio separability, spatial stability, zero-shot anatomical grounding, and cross-attention localization).

\paragraph{Contributions.}
\begin{itemize}
    \item A bilateral slot-based architecture where unsupervised slots serve as correspondence units and bidirectional cross-attention enables structured cross-eye information exchange.
    \item A statistically robust evaluation (\(n=10\) seeds, 95\% CIs, Wilcoxon test) under matched-backbone settings, establishing a 4.2-point AUC gain over a fully fine-tuned RETFound ViT-L baseline.
    \item Mechanistic analyses---pairing disruption, stress testing, representational separability, spatial stability, zero-shot grounding, and cross-attention localization---that collectively characterize correspondence-dependent behavior.
\end{itemize}

\section{Related Works}

\subsection{Bilateral Retinal Analysis}
Bilateral retinal analysis has been explored through Siamese CNNs with shared weights and late fusion for diabetic retinopathy grading and glaucoma assessment~\cite{ref_bilateral1,ref_bilateral2,ref_bilateral3}. In OCT, inter-eye asymmetry analysis has proven valuable for early glaucoma detection~\cite{ref_oct1}. Recent work incorporates attention mechanisms and cross-eye feature alignment constraints toward structured comparison. However, these approaches share a critical limitation: they operate on global or grid-based features without explicitly modeling \emph{which} structures correspond across eyes. Without correspondence units, it is difficult to disentangle genuine anatomical comparison from learned statistical associations. We introduce object-centric slots as explicit correspondence units, enabling direct measurement of cross-eye anatomical alignment.

\subsection{Medical Foundation Models}
Self-supervised pretraining has reshaped medical image analysis. Methods such as MAE~\cite{ref_mae}, DINO/DINOv2~\cite{ref_dino,ref_dinov2}, and BYOL~\cite{ref_byol} with ViT~\cite{ref_vit} and Swin~\cite{ref_swin} backbones yield strong representations. RETFound~\cite{ref_retfound} applies MAE-style pretraining to fundus and OCT images, achieving state-of-the-art ocular and systemic disease prediction. We build on RETFound, introducing a specialized bilateral module to isolate structured correspondence from general representation quality.

\subsection{Object-Centric Learning}
Object-centric learning decomposes scenes into constituent entities without supervision. Early approaches (IODINE~\cite{ref_iodine}, MONet~\cite{ref_monet}) used iterative variational inference on synthetic data. Slot Attention~\cite{ref_slot} introduced scalable competitive attention where slot vectors compete to explain input features through iterative soft-clustering, scaling to complex natural images. Within medical imaging, object-centric methods remain largely unexplored despite the natural alignment between anatomical structures and object-like entities---optic disc, macula, and vessel arcades are precisely the compositional entities slot-based methods discover. We are among the first to apply slot decomposition to medical images and the first to integrate slots with cross-view correspondence for bilateral reasoning. We treat slots as soft correspondence units whose alignment can be measured, not as perfect segmentations.

\section{Method}

\subsection{Overall Architecture}
\begin{figure}[htbp]
\centering
\includegraphics[width=1.0\linewidth]{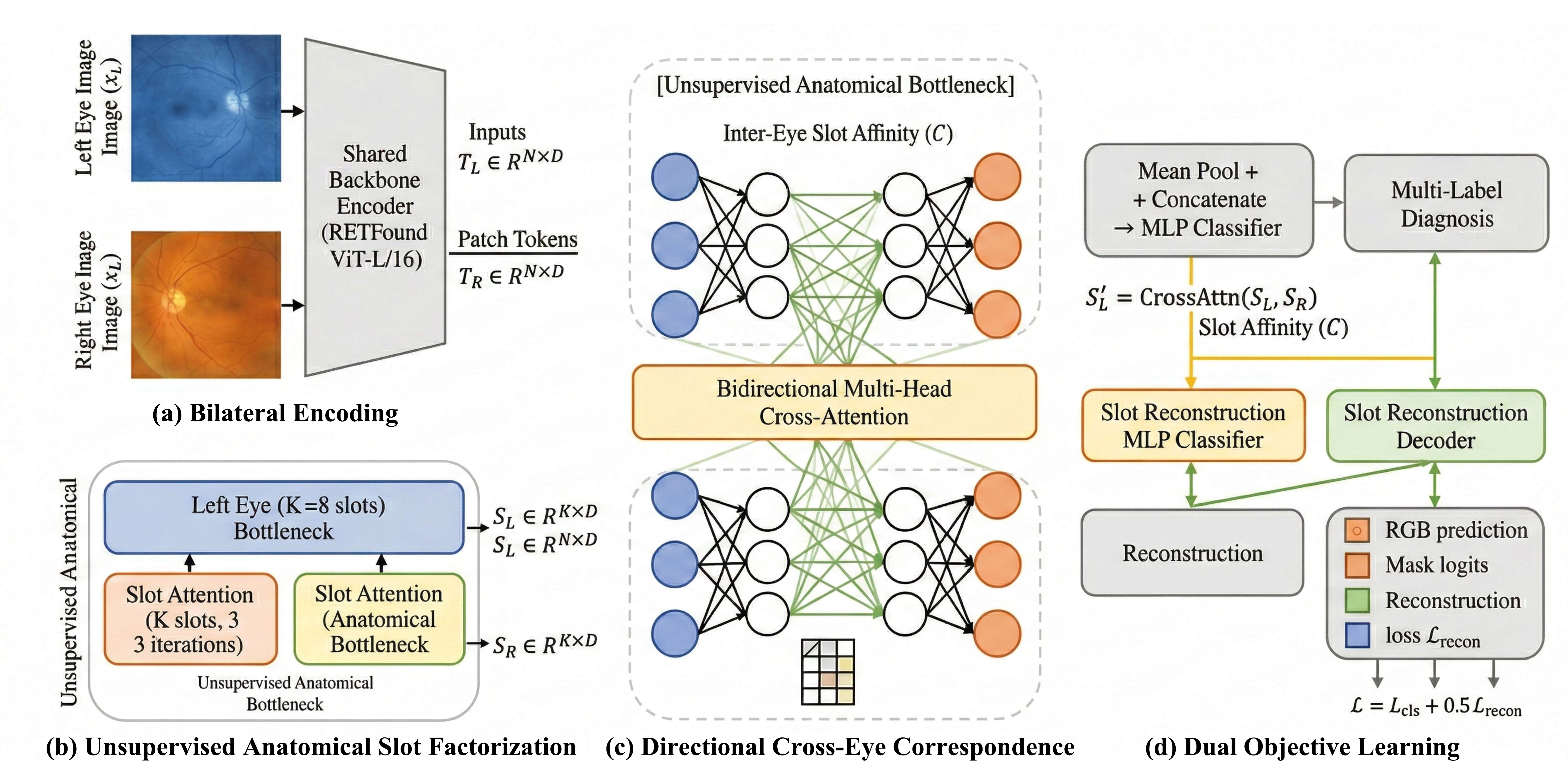}
\caption{Anatomy-Slot pipeline. A bilateral pair is encoded by a shared ViT backbone; Slot Attention yields $K$ slots per eye. Bidirectional cross-attention aligns homologous slots, pooled features are concatenated for diagnosis, and a lightweight decoder reconstructs low-resolution RGB to stabilize slot learning.}
\label{fig:pipeline}
\end{figure}

Given bilateral fundus images $(x_L, x_R)$, a shared RETFound ViT-L/16 backbone~\cite{ref_retfound} extracts patch tokens $T_L, T_R \in \mathbb{R}^{N \times D}$ from the final transformer layer ($N{=}196$, $D{=}1024$ for $224{\times}224$ inputs). Slot Attention maps these into $K$ slot vectors per eye ($S_L, S_R \in \mathbb{R}^{K \times D}$). Bidirectional cross-attention refines slots by attending to the contralateral eye, producing $S_L', S_R'$. Per-eye features are mean-pooled, concatenated into $z = [\bar{s}_L; \bar{s}_R]$, and classified via a two-layer MLP (hidden dim.\ 512, GELU, dropout 0.1). A lightweight decoder reconstructs low-resolution RGB patches from slots to prevent collapse. The architecture is modular: backbone, slot module, cross-attention, and decoder can be independently ablated (Section~\ref{sec:mech}). Figure~\ref{fig:pipeline} illustrates the overall pipeline.

\subsection{Slot Attention Module}
For each eye, patch tokens $T \in \mathbb{R}^{N\times D}$ are mapped into $K$ slots via iterative competitive attention ($I$ refinement steps). Slots are initialized as $s_i^{0} = \mu_i + \sigma_i \odot \epsilon$ with learnable $\mu_i, \sigma_i \in \mathbb{R}^D$ and $\epsilon \sim \mathcal{N}(0, \mathbf{I})$. At iteration $t$, layer normalization is applied to both inputs and slots before attention:

\begin{align}
q_i &= W_q\, \mathrm{LN}(s_i^{t-1}), \quad
k_j = W_k\, \mathrm{LN}(t_j), \quad
v_j = W_v\, \mathrm{LN}(t_j), \\
a_{ij} &= \mathrm{softmax}_i\!\left(\frac{q_i^\top k_j}{\sqrt{D}}\right), \quad
\tilde{a}_{ij} = \frac{a_{ij}}{\sum_j a_{ij}},\\
u_i &= \sum\nolimits_j \tilde{a}_{ij}\, v_j, \quad
s_i^{t} = \mathrm{GRU}(u_i,\; s_i^{t-1}),
\end{align}
where $W_q, W_k, W_v \in \mathbb{R}^{D \times D}$ are shared across iterations. The softmax over slots (dimension $i$) implements competition: slots vie to explain each patch. Attention weights are then renormalized over tokens (dimension $j$), yielding soft, multi-modal assignment masks without explicit sparsity constraints. Backbone positional embeddings are retained throughout. Our renormalization applies L1 normalization over tokens ($\sum_j a_{ij}$), following the standard Slot Attention formulation~\cite{ref_slot}. The number of slots $K$ is a hyperparameter determined via grid search (see Section~\ref{sec:impl}); we use $K=8$ by default.

\subsection{Bilateral Cross-Attention}
Cross-eye correspondence is established through multi-head cross-attention ($H=8$ heads):

\begin{align}
S_L' &= \mathrm{MultiHeadCrossAttn}(Q=S_LW_Q^c,\; K=S_RW_K^c,\; V=S_RW_V^c), \\
S_R' &= \mathrm{MultiHeadCrossAttn}(Q=S_RW_Q^c,\; K=S_LW_K^c,\; V=S_LW_V^c),
\end{align}
where $W_Q^c, W_K^c, W_V^c \in \mathbb{R}^{D \times D_h}$ project to per-head dimension $D_h = D/H$. Multiple heads allow distinct correspondence patterns (e.g., vessel alignment vs.\ optic disc symmetry).

\subsection{Reconstruction Objective}
To prevent slot collapse, each refined slot predicts per-patch RGB and assignment logits via a shared linear decoder:
\begin{equation}
(r_{k,n},\; m_{k,n}) = \mathrm{Decoder}(s_k',\; p_n),
\end{equation}
where $p_n$ is a learned patch positional encoding. Reconstructed patches are a soft combination:
\begin{equation}
\hat{x}_n = \sum_{k=1}^{K} \pi_{k,n} \cdot \sigma(r_{k,n}), \quad
\pi_{k,n} = \mathrm{softmax}_k(m_{k,n}).
\end{equation}
The reconstruction loss is $\mathcal{L}_{recon} = \frac{1}{N}\sum_{n=1}^{N}\|\hat{x}_n - x_n\|_2^2$, where $x_n$ is the downsampled ground-truth RGB patch. The classification loss $\mathcal{L}_{cls}$ is the binary cross-entropy over $C=8$ diagnostic categories:
\begin{equation}
\mathcal{L}_{cls} = -\frac{1}{C}\sum_{c=1}^{C}\Big[y_c\log(\sigma(z_c)) + (1-y_c)\log(1-\sigma(z_c))\Big],
\end{equation}
where $z_c$ is the output logit for class $c$, $\sigma(\cdot)$ is the sigmoid function, and $y_c \in \{0,1\}$ is the ground-truth label. The total objective is $\mathcal{L} = \mathcal{L}_{cls} + 0.5\,\mathcal{L}_{recon}$ ($\lambda=0.5$ from grid search over $\{0.1, 0.25, 0.5, 1.0\}$, which yielded validation AUC of $0.839$, $0.853$, $0.861$, and $0.847$, respectively).

\subsection{Implementation Details}
\label{sec:impl}

\paragraph{Pretraining.} We pretrain the slot module and decoder on \textasciitilde{}14k unlabeled retinal images pooled from publicly available datasets (ODIR training split~\cite{ref_odir}, APTOS2019~\cite{ref_aptos2019}, IDRiD~\cite{ref_idrid}, MESSIDOR2~\cite{ref_messidor2}, PAPILA~\cite{ref_papila}, Retina~\cite{ref_retina}); all labels are discarded during pretraining. To prevent data leakage, images from the ODIR test split are excluded from the pretraining pool. The RETFound backbone is frozen except for the final transformer block (block 23) and layer normalization parameters. Slot Attention: $I=5$ iterations, $K=8$ slots, selected via grid search over $\{2, 4, 6, 8, 12, 16\}$. AdamW optimizer: learning rate $3 \times 10^{-4}$, weight decay 0.05, batch size 32 bilateral pairs, 100 epochs with cosine annealing (5 warmup epochs), using only $\mathcal{L}_{recon}$.

\paragraph{Main training.} On ODIR-5K~\cite{ref_odir}, we initialize from RETFound and load pretrained slot weights, unfreezing the same parameter subset. We follow the standard 70/15/15 train/val/test split. Slot Attention: $I=3$ iterations, $K=8$ slots. AdamW: learning rates $1 \times 10^{-4}$ (backbone) and $5 \times 10^{-4}$ (slot module and classifier), cosine schedule over 50 epochs (3 warmup epochs). Binary cross-entropy for $\mathcal{L}_{cls}$. Images resized to $224 \times 224$. Data augmentation during training consists of random horizontal flips, random rotation ($\pm 15^\circ$), and brightness/contrast jitter ($\pm 10\%$); left and right eyes receive identical transforms to preserve bilateral correspondence. Mixed-precision (FP16) training on a single NVIDIA A100-80GB completes in approximately 2.5 hours per run.

\paragraph{Computational complexity.} The RETFound ViT-L backbone has 307M parameters and 61.6 GFLOPs at $224{\times}224$. The Slot Attention module (Q/K/V projections, GRU) adds 9.4M; bidirectional cross-attention (Q/K/V, 8 heads) adds 4.2M; the reconstruction decoder adds 0.1M. Total overhead is 13.7M (4.5\% of backbone). Slot and cross-attention over $K{=}8$, $N{=}196$ incur negligible FLOPs ($<$0.1 GFLOPs). Inference latency increases $\sim$3\% vs.\ backbone-only on an A100-80GB.

\subsection{Correspondence Structure Analysis}
Let $A^{h}\in\mathbb{R}^{K\times K}$ denote the cross-eye slot affinity matrix for head $h$, where $A^{h}_{ij}$ measures attention from left-eye slot $i$ to right-eye slot $j$. The mean correspondence matrix $C = \frac{1}{H}\sum_{h=1}^H A^{h}$ yields the diagonal concentration ratio:
\begin{equation}
\rho = \frac{\sum_i C_{ii}}{\sum_{i,j} C_{ij}} \in [0, 1],
\end{equation}
where $\rho \approx 1/K$ indicates uniform cross-attention, while $\rho \rightarrow 1$ indicates each left-eye slot exclusively attends to a single homologous right-eye slot. Beyond diagonal concentration, we report the mean off-diagonal row maximum $\bar{m}_{\text{off}}=0.08$ (confirming that each slot has at most one weak secondary counterpart) and the mean per-slot attention entropy $H=1.2$ bits (chance: $\log_2 8 = 3$ bits), both indicating concentrated rather than diffuse cross-attention.

\section{Experiments and Analyses}

Our experiments are structured around the two research questions posed in Section~1: we first evaluate diagnostic accuracy and ablations (RQ1), followed by mechanistic and interpretability analyses designed to probe correspondence-dependent behavior (RQ2).

\subsection{Experimental Setup}
\paragraph{Datasets.} ODIR-5K~\cite{ref_odir} (5,000 patients, bilateral fundus photographs, multi-label: Normal, Diabetes, Glaucoma, Cataract, AMD, Hypertension, Myopia, Other). Pretraining pool: ODIR, APTOS2019, IDRiD, MESSIDOR2, PAPILA, Retina (\textasciitilde{}14k images). External validation (DR only): EyePACS~\cite{ref_eyepacs}. Anatomical grounding: REFUGE validation set~\cite{ref_refuge} (400 images, pixel-level optic disc/cup annotations).

\paragraph{Backbone.} RETFound ViT-L/16 (307M parameters)~\cite{ref_retfound}. See Section~\ref{sec:impl} for full optimization details. Default $K=8$ slots; swept $K \in \{2, 4, 6, 8, 12, 16\}$.

\paragraph{Statistics.} $n=10$ independent seeds per configuration. Metrics: macro-averaged AUC and F1, reported as $\mu \pm \sigma$ with 95\% CIs. Shapiro--Wilk tests rejected normality for several configurations; all paired comparisons use Wilcoxon signed-rank test (exact $W$, two-sided $p$).

\subsection{Main Results, Ablations, and Capacity}

\begin{table}[htbp]
\caption{Main results and ablation on ODIR-5K ($n=10$). Wilcoxon signed-rank test vs.\ baseline: $W=0$, $p=0.002$.}\label{tab:main}
\centering
\footnotesize
\setlength{\tabcolsep}{3.5pt}
\begin{tabular}{@{}lcccc@{}}
\toprule
\textbf{Model} & \textbf{F1} & \textbf{AUC} & \textbf{95\% CI} & \textbf{Config.} \\ \midrule
Baseline (RETFound-L) & $0.481_{\pm 0.007}$ & $0.823_{\pm 0.005}$ & $[0.819, 0.827]$ & Full fine-tuning \\ \midrule
\textit{Ablation Study:} & & & & \\
~~Frozen backbone       & $0.452_{\pm 0.008}$ & $0.785_{\pm 0.009}$ & $[0.778, 0.792]$ & Frozen BB \\
~~w/o $\mathcal{L}_{recon}$ & $0.459_{\pm 0.008}$ & $0.729_{\pm 0.006}$ & $[0.724, 0.734]$ & Slot collapse \\
~~PatchCrossAttn        & $0.518_{\pm 0.007}$ & $0.837_{\pm 0.006}$ & $[0.832, 0.842]$ & Patch X-attn \\
~~w/o Slots             & $0.515_{\pm 0.005}$ & $0.828_{\pm 0.004}$ & $[0.825, 0.831]$ & Global pooling \\
~~w/o Slots + $\mathcal{L}_{recon}$ & $0.522_{\pm 0.006}$ & $0.834_{\pm 0.005}$ & $[0.830, 0.838]$ & Pool + recon. \\
~~w/o Bilateral         & $0.462_{\pm 0.009}$ & $0.754_{\pm 0.007}$ & $[0.749, 0.759]$ & No cross-eye \\
\midrule
\textbf{Anatomy-Slot}   & $\mathbf{0.535_{\pm 0.006}}$ & $\mathbf{0.865_{\pm 0.004}}$ & $[\mathbf{0.862, 0.868}]$ & \textbf{Full ($K{=}8$)} \\ \bottomrule
\end{tabular}
\end{table}

Table~\ref{tab:main} summarizes the main results. The RETFound ViT-L baseline achieves 0.823 AUC (F1 0.481). Anatomy-Slot improves to 0.865 AUC (F1 0.535), a 4.2-point gain ($W{=}0$, $p{=}0.002$; Anatomy-Slot wins all 10 seeds).

The Baseline concatenates \texttt{[CLS]} tokens from both eyes; \emph{w/o Slots} applies global average pooling to all patch tokens before concatenation. Their minimal difference (0.828 vs.\ 0.823) indicates pooling alone provides no meaningful correspondence, while the gap to Anatomy-Slot (0.865) underscores the role of slot-based alignment.

The ablation pattern reveals an interaction effect. Bilateral processing without slots (\emph{w/o Slots}) yields marginal gain (0.828 vs.\ 0.823), while slots without cross-eye interaction (\emph{w/o Bilateral}) underperform (0.754). Removing reconstruction (\emph{w/o $\mathcal{L}_{recon}$}) causes severe degradation (0.729), consistent with slot collapse. Freezing the backbone (0.785) underperforms. PatchCrossAttn (0.837) outperforms global pooling but substantially trails Anatomy-Slot (0.865), validating that the slot bottleneck is essential. Adding reconstruction to the no-slot baseline (0.834, $p{=}0.002$) confirms structured slots contribute beyond reconstruction alone. Wilcoxon tests confirm PatchCrossAttn ($p{=}0.002$) and w/o Slots ($p{=}0.002$) are significantly inferior.

\paragraph{Per-class bilateral analysis.}
Bilateral information is not uniformly informative across categories. Glaucoma assessment relies heavily on inter-eye cup-to-disc ratio asymmetry, while cataract and myopia are diagnosed primarily from monocular features. Table~\ref{tab:perclass} reports per-class AUC. The largest gains occur for Glaucoma ($+9$) and Diabetic Retinopathy ($+8$)---conditions where inter-eye asymmetry is diagnostic---while monocular conditions (Cataract $+1$, Myopia $+1$) show smaller improvements. This supports that cross-eye correspondence benefits diagnosis proportionally to inherent inter-eye asymmetry.

\begin{table}[htbp]
\caption{Per-class AUC on ODIR-5K ($n=10$ mean). Per-class std.\ ranges from $\pm 0.01$ to $\pm 0.02$ across categories. N: Normal, D: Diabetes, G: Glaucoma, C: Cataract, A: AMD, H: Hypertension, M: Myopia, O: Other.}\label{tab:perclass}
\centering
\small
\setlength{\tabcolsep}{3pt}
\begin{tabular}{@{}lcccccccc@{}}
\toprule
\textbf{Model} & \textbf{N} & \textbf{D} & \textbf{G} & \textbf{C} & \textbf{A} & \textbf{H} & \textbf{M} & \textbf{O} \\ \midrule
Baseline (RETFound-L) & 0.84 & 0.83 & 0.79 & 0.81 & 0.81 & 0.82 & 0.85 & 0.83 \\
Anatomy-Slot (Ours)   & 0.87 & 0.91 & 0.88 & 0.82 & 0.88 & 0.84 & 0.86 & 0.86 \\
\midrule
$\Delta$ (Ours $-$ Baseline) & +3 & +8 & +9 & +1 & +7 & +2 & +1 & +3 \\
\bottomrule
\end{tabular}
\end{table}

\begin{figure}[t]
\centering
\includegraphics[width=1.0\linewidth]{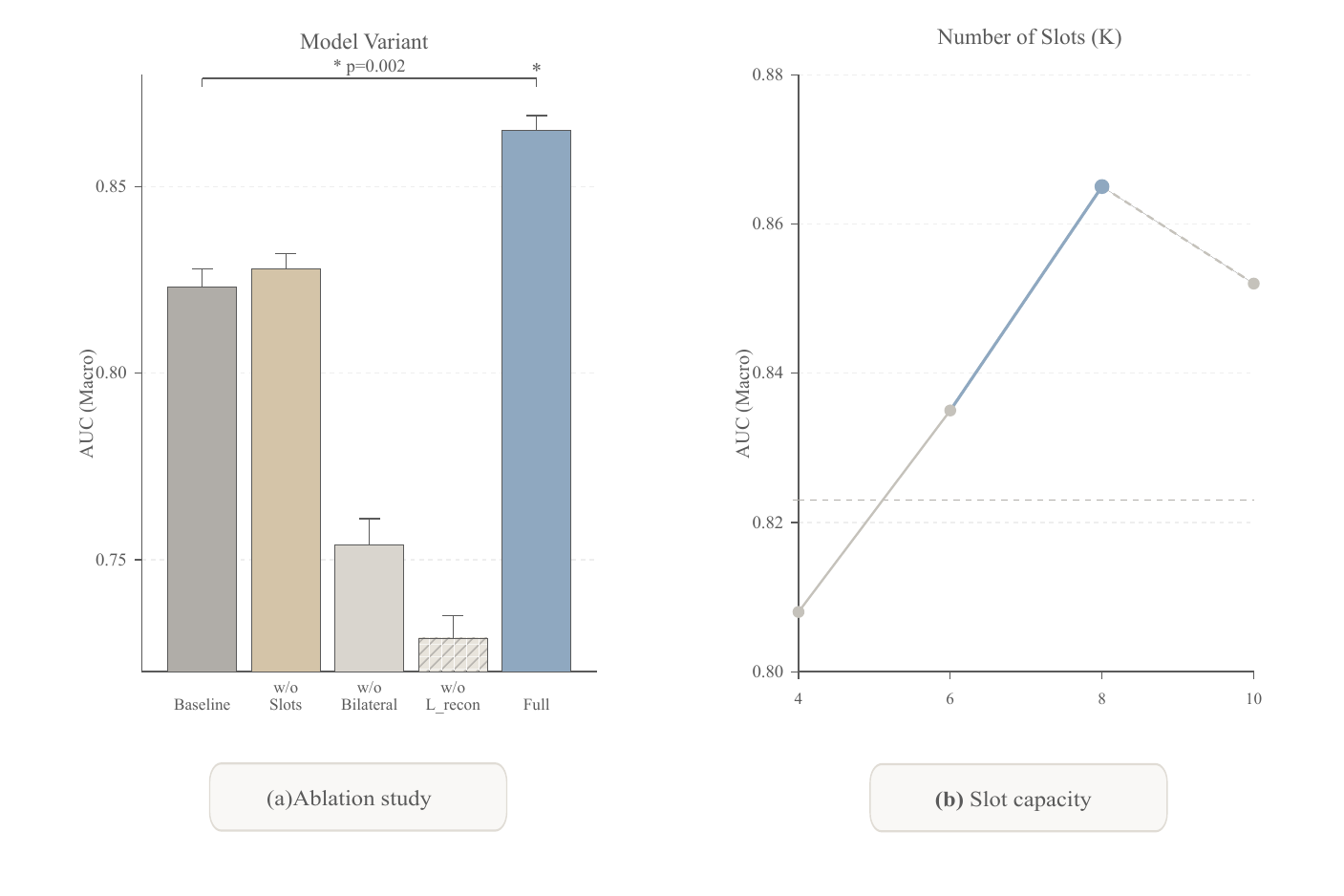}
\caption{Architecture ablation and capacity analysis on ODIR-5K (AUC macro). (a) Ablation: baseline, bilateral-only, slots-only, no-reconstruction, full model. (b) Slot capacity sweep: performance peaks at $K=8$. Error bars: $\pm$1 s.d.\ ($n=10$); $^{**}$: $p=0.002$ (Wilcoxon vs.\ baseline).}
\label{fig:slot_capacity}
\end{figure}

\paragraph{Slot capacity.} Figure~\ref{fig:slot_capacity}(b) shows validation AUC vs.\ $K$. Performance rises from 0.816 ($K{=}2$) through 0.838 ($K{=}4$) and 0.855 ($K{=}6$) to a peak of 0.861 ($K{=}8$), then declines at 0.850 ($K{=}12$) and 0.834 ($K{=}16$). The rise reflects increasing capacity to represent anatomical regions ($K{=}2$ captures foreground/background; $K{=}6$ separates disc, macula, vessels). The decline reflects correspondence dilution: excess slots disperse cross-attention, weakening the correspondence signal. Optimal $K{=}8$ matches the number of visually distinguishable anatomical zones in a typical fundus. For smaller or unseen datasets, we recommend $K{\in}[4,8]$.

\subsection{Mechanistic Tests: Pairing Disruption and Robustness}
\label{sec:mech}

\begin{table}[htbp]
\caption{Mechanistic evaluation on ODIR-5K ($n{=}10$ mean AUC). Std.\ ranges $\pm 0.004$ (clean) to $\pm 0.022$ ($\sigma{=}0.2$).}\label{tab:mech}
\centering
\footnotesize
\setlength{\tabcolsep}{3pt}
\begin{minipage}{0.45\linewidth}
\centering
\textit{Pairing disruption}\\[2pt]
\begin{tabular}{@{}lcc@{}}
\toprule
Experiment & Ours & Baseline \\
\midrule
Standard       & \textbf{0.865} & 0.823 \\
Shuffle eval.  & 0.674 & 0.791 \\
Shuffled-pair  & 0.721 & 0.796 \\
\bottomrule
\end{tabular}
\end{minipage}\hfill
\begin{minipage}{0.52\linewidth}
\centering
\textit{Stress testing under $\mathcal{N}(0,\sigma^2)$}\\[2pt]
\begin{tabular}{@{}lcccc@{}}
\toprule
Model & $\sigma{=}0$ & $\sigma{=}0.05$ & $\sigma{=}0.1$ & $\sigma{=}0.2$ \\
\midrule
Baseline    & 0.823 & 0.718 & 0.641 & 0.572 \\
w/o Slots   & 0.828 & 0.742 & 0.667 & 0.591 \\
\textbf{Ours}& \textbf{0.865} & \textbf{0.809} & \textbf{0.761} & \textbf{0.682} \\
\bottomrule
\end{tabular}
\end{minipage}
\end{table}

\paragraph{Pairing disruption.} If Anatomy-Slot relies on cross-eye correspondence, disrupting pairing should degrade performance. \emph{Shuffle at evaluation} (test-time random left--right reassignment) drops Anatomy-Slot from 0.865 to 0.674 AUC (19.1-point drop) vs.\ 3.2 points for the baseline (0.823 $\rightarrow$ 0.791). \emph{Shuffled-pair training} (training with randomly paired images from different patients) yields 0.721, confirming disrupted pairing substantially impairs correspondence learning. The asymmetric drop supports correspondence-dependent behavior; the baseline's small drop suggests its bilateral gain reflects statistical co-occurrence rather than structural comparison.

\paragraph{Stress testing.} Table~\ref{tab:mech} (right) reports Gaussian noise robustness. Anatomy-Slot is strictly better than both baselines at all levels. At $\sigma{=}0.2$, it retains 0.682 vs.\ 0.572 (baseline) and 0.591 (no-slots). The gap widens with noise---from 4.2 points (clean) to 9.1--11.0 ($\sigma{=}0.2$). We hypothesize slot-based decomposition isolates noise: perturbations affect individual slots rather than corrupting the global representation, and cross-attention can down-weight noisy slots during matching. The monotonic gap--noise relationship is consistent with this interpretation.

\subsection{Representation Analysis and Anatomical Grounding}

\paragraph{Representational separability.} Fisher Ratio $\mathrm{Tr}(S_W^{-1} S_B)$ in the penultimate space: 0.85 (baseline), 0.98 (no-slots), 1.62 (Anatomy-Slot). All models produce matched-dimension vectors ($2D$), ruling out dimensionality artifacts. Improved separability is consistent with slot-conditioned features aligning by anatomical rather than textural criteria.

\paragraph{Spatial stability.} Mean per-slot spatial variance across 3,500 bilateral pairs is $4.12$ ($14{\times}14$ grid), indicating consistent anatomical targeting. We further compute a topological consistency score---Jaccard index of dominant-attention patch sets ($\mathrm{argmax}$) per slot across all pairs---yielding $0.78{\pm}0.09$. This confirms slot semantics are globally consistent: Slots~1--2 center on optic disc and macula, Slots~3--5 on vascular arcades, Slots~6--8 on peripheral retina (Fig.~\ref{fig:slot}(a)).

\paragraph{Zero-shot anatomical grounding.} We evaluate alignment between slot assignments and expert optic disc annotations on REFUGE~\cite{ref_refuge} (400 images) without segmentation fine-tuning. Binary slot masks via $\mathrm{argmax}$ over slot assignment probabilities at each patch; Slot 1 (highest optic disc overlap) is evaluated.

\begin{table}[htbp]
\caption{Optic disc grounding on REFUGE (400 images). Zero-shot; masks via $\mathrm{argmax}$, no threshold tuning. Mean $\pm$ std.\ ($n{=}10$).}\label{tab:refuge}
\centering
\footnotesize
\begin{tabular}{@{}lcc@{}}
\toprule
\textbf{Method} & \textbf{mIoU} & \textbf{Dice} \\
\midrule
K-Means & $0.452_{\pm 0.110}$ & $0.623$ \\
Anatomy-Slot (Slot~1) & $0.635_{\pm 0.015}$ & $0.777$ \\
\bottomrule
\end{tabular}
\end{table}

Anatomy-Slot achieves mIoU 0.635 (Dice 0.777) on REFUGE, substantially exceeding K-Means (mIoU 0.452, Dice 0.623; Table~\ref{tab:refuge}). The lower std.\ (0.015 vs.\ 0.110) indicates consistent alignment. Chance-level mIoU is $<$0.01. Slot~1 was identified post-hoc as the slot with highest optic disc overlap; mean mIoU across all $K{=}8$ slots is $0.124{\pm}0.080$, confirming non-uniform specialization---only Slot~1 targets the disc. This zero-shot grounding, emerging from reconstruction-driven attention without segmentation labels, is notable but partial (Dice 0.777 vs.\ 1.0).

\subsection{Cross-Attention Localization and Qualitative Analysis}

\begin{figure}[ht]
\centering
\includegraphics[width=1.0\linewidth]{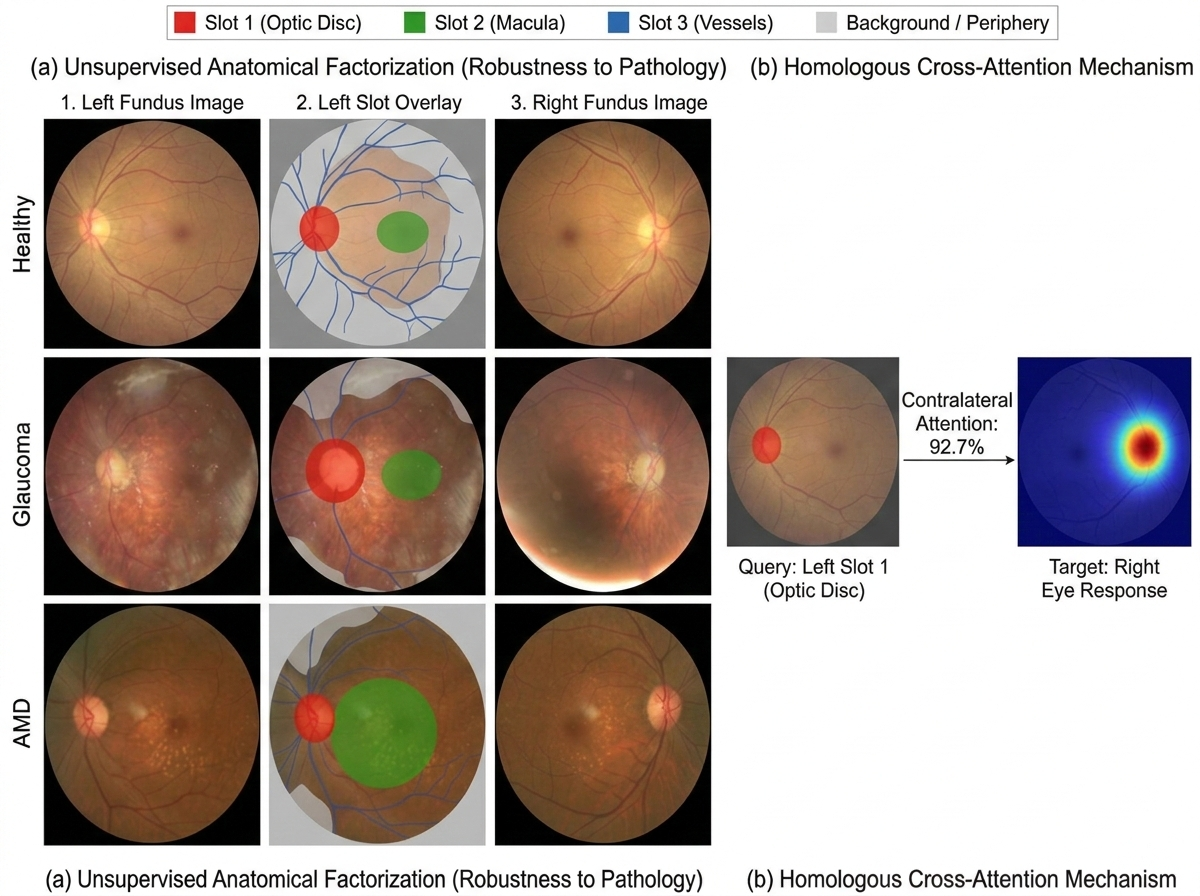}
\caption{(a) Emergent anatomical alignment across three ODIR cases (healthy, glaucoma, AMD). Slot 1 (red) centers on optic disc, Slot 2 (green) on macula, Slot 3 (blue) on vessel arcades, remaining slots (gray) cover periphery. Right-eye images shown for bilateral context. (b) Homologous cross-attention: left-eye optic disc slot queries the right eye and concentrates on the contralateral optic disc (mean contralateral attention across all pairs: 92.7\%).}
\label{fig:slot}
\end{figure}

Figure~\ref{fig:slot}(a) visualizes slot assignments across healthy, glaucomatous, and AMD-affected retinas. Slot assignments are largely disease-invariant: Slot~1 consistently targets the optic disc. Boundaries are soft, reflecting continuous anatomy; peripapillary atrophy is captured jointly by Slot~1 and peripheral slots.

Figure~\ref{fig:slot}(b) shows cross-attention for a left-eye optic disc slot querying the right eye: 92.7\% of attention mass concentrates within $3{\times}3$ patches of the contralateral disc center. The diagonal concentration ratio $\rho{=}0.71$ ($\rho_{random}{=}0.125$ for $K{=}8$) confirms structured, non-random correspondence.

\subsection{Benchmarking}

\begin{table}[htbp]
\caption{Comparison with recent methods on ODIR-5K (AUC, $224{\times}224$ input). Rows below the midrule are implemented by the authors under matched protocols (MedMamba-Bilateral adapts~\cite{ref_medmamba} with bilateral fusion). Values are single-run except Anatomy-Slot ($n{=}10$).}\label{tab:bench}
\centering
\footnotesize
\setlength{\tabcolsep}{4pt}
\begin{tabular}{@{}llll@{}}
\toprule
Category & Model & Pretraining & AUC \\
\midrule
General FM  & DINOv2-ViT-L/14~\cite{ref_dinov2}  & LVD-142M (SSL)    & 0.782 \\
General FM  & DINOv3-ViT-L/16~\cite{ref_dinov3}  & Web-scale (SSL)   & 0.801 \\
Medical FM  & RETFound-MAE~\cite{ref_retfound}    & Retinal (SSL)     & 0.823 \\
\midrule
Mamba       & MedMamba w/ bilateral fusion~\cite{ref_medmamba} & Labels   & 0.832 \\
Bilateral   & Swin-B w/ cross-eye attn.           & Labels            & 0.841 \\
Anatomy-Aware & ViT-L w/ landmark supervision     & Labels + masks    & 0.845 \\
\midrule
\textbf{Ours} & \textbf{Anatomy-Slot}             & \textbf{Slots + labels} & \textbf{0.865} \\
\bottomrule
\end{tabular}
\end{table}

Table~\ref{tab:bench} contextualizes Anatomy-Slot. General vision FMs (DINOv2, DINOv3) underperform medical-specific pretraining (RETFound). Methods incorporating bilateral fusion or anatomical priors improve over RETFound, supporting structured bilateral modeling. Anatomy-Slot outperforms all methods (0.865) using only image-level labels and unsupervised slots; the strongest non-slot baseline requires pixel-level annotations for comparable performance.

\subsection{External Validation}
We evaluate diabetic retinopathy (DR) classification on EyePACS~\cite{ref_eyepacs} under direct transfer, since EyePACS provides DR labels only. On ODIR, DR-specific AUC is 0.83 (baseline) vs.\ 0.91 (ours), a $+8$-point gap (Table~\ref{tab:perclass}). On EyePACS, baseline drops to 0.782 and Anatomy-Slot to 0.842, yielding a $+6.0$-point gap. The structured bilateral advantage persists under distribution shift, though the DR-specific gap narrows from 8.0 to 6.0 points. We note slot semantics on EyePACS are unverified; we refrain from attributing the reduced gap magnitude solely to structural priors.

\section{Discussion and Conclusion}
We introduced Anatomy-Slot, decomposing fundus images into unsupervised slots aligned via bidirectional cross-attention. The core insight: slot decomposition provides a natural bottleneck for anatomical correspondence---unlike direct cross-attention over patches, slots specialize to coherent anatomical regions, enabling structured, measurable cross-eye alignment ($\rho{=}0.71$). Under matched-backbone settings ($n{=}10$), Anatomy-Slot achieves a 4.2-point AUC gain over fine-tuned RETFound ViT-L ($p{=}0.002$). Ablations confirm the gain requires slot decomposition plus bilateral cross-attention. Mechanistic tests reveal correspondence-dependent behavior: 19.1-point drop under pairing disruption, superior noise robustness. Slots yield emergent anatomical alignment (optic disc Dice 0.777, zero-shot).

\paragraph{Limitations.} Primary evaluation is on ODIR-5K; multi-center validation across diverse populations and imaging devices is needed to assess generalization. Anatomical coverage is partially verified---quantitative for optic disc, qualitative for macula and vessels. Systematic expert annotation of slot assignments across all landmarks would strengthen anatomy-aware correspondence claims. Tolerance to clinically realistic misalignment (strabismus, monocular patients, poorly registered pairs) and severe localized pathology (large hemorrhages, tumors, dense exudates) remains untested. Finally, establishing correspondence-based reasoning as the definitive causal mechanism requires convergent evidence; our suite of mutually reinforcing analyses---disruption, separability, stability, grounding, and cross-attention---forms a coherent behavioral signature toward this goal.

\paragraph{Future work.} Immediate extensions include multi-center validation, systematic ophthalmologist-annotated landmark verification, and testing under realistic pairing failure modes. The framework extends to other paired-organ contexts (mammography, chest radiography, brain MRI). Methodologically, explicit symmetry constraints in cross-attention and hierarchical slot decompositions are natural next steps. Causal mediation analyses and counterfactual manipulations can further isolate correspondence contributions to clinical decisions.

\bibliographystyle{splncs04}
\bibliography{reference}

@inproceedings{ref_mae,
  title     = {Masked Autoencoders Are Scalable Vision Learners},
  author    = {He, Kaiming and Chen, Xinlei and Xie, Saining and Li, Yanghao and Doll{\'a}r, Piotr and Girshick, Ross},
  booktitle = {Proceedings of the IEEE/CVF Conference on Computer Vision and Pattern Recognition (CVPR)},
  pages     = {16000--16009},
  year      = {2022}
}

@inproceedings{ref_dino,
  title     = {Emerging Properties in Self-Supervised Vision Transformers},
  author    = {Caron, Mathilde and Touvron, Hugo and Misra, Ishan and J{\'e}gou, Herv{\'e} and Mairal, Julien and Bojanowski, Piotr and Joulin, Armand},
  booktitle = {Proceedings of the IEEE/CVF International Conference on Computer Vision (ICCV)},
  pages     = {9650--9660},
  year      = {2021}
}

@article{ref_byol,
  title={Bootstrap your own latent-a new approach to self-supervised learning},
  author={Grill, Jean-Bastien and Strub, Florian and Altch{\'e}, Florent and Tallec, Corentin and Richemond, Pierre and Buchatskaya, Elena and Doersch, Carl and Avila Pires, Bernardo and Guo, Zhaohan and Gheshlaghi Azar, Mohammad and others},
  journal={Advances in neural information processing systems},
  volume={33},
  pages={21271--21284},
  year={2020}
}

@article{ref_vit,
  title={An image is worth 16x16 words: Transformers for image recognition at scale},
  author={Dosovitskiy, Alexey and Beyer, Lucas and Kolesnikov, Alexander and Weissenborn, Dirk and Zhai, Xiaohua and Unterthiner, Thomas and Dehghani, Mostafa and Minderer, Matthias and Heigold, Georg and Gelly, Sylvain and others},
  journal={arXiv preprint arXiv:2010.11929},
  year={2020}
}

@inproceedings{ref_swin,
  title     = {Swin Transformer: Hierarchical Vision Transformer Using Shifted Windows},
  author    = {Liu, Ze and Lin, Yutong and Cao, Yue and Hu, Han and Wei, Yixuan and Zhang, Zheng and Lin, Stephen and Guo, Baining},
  booktitle = {Proceedings of the IEEE/CVF International Conference on Computer Vision (ICCV)},
  pages     = {10012--10022},
  year      = {2021}
}

@inproceedings{ref_iodine,
  title={Multi-object representation learning with iterative variational inference},
  author={Greff, Klaus and Kaufman, Rapha{\"e}l Lopez and Kabra, Rishabh and Watters, Nick and Burgess, Christopher and Zoran, Daniel and Matthey, Loic and Botvinick, Matthew and Lerchner, Alexander},
  booktitle={International conference on machine learning},
  pages={2424--2433},
  year={2019},
  organization={PMLR}
}

@misc{ref_monet,
  title     = {MONet: Unsupervised Scene Decomposition and Representation},
  author    = {Burgess, Christopher P and Matthey, Loic and Watters, Nicholas and Kabra, Rishabh and Higgins, Irina and Botvinick, Matt and Lerchner, Alexander},
  howpublished = {arXiv preprint arXiv:1901.11390},
  year      = {2019}
}

@article{ref_slot,
  title={Object-centric learning with slot attention},
  author={Locatello, Francesco and Weissenborn, Dirk and Unterthiner, Thomas and Mahendran, Aravindh and Heigold, Georg and Uszkoreit, Jakob and Dosovitskiy, Alexey and Kipf, Thomas},
  journal={Advances in neural information processing systems},
  volume={33},
  pages={11525--11538},
  year={2020}
}

@article{ref_dinov2,
  title={Dinov2: Learning robust visual features without supervision},
  author={Oquab, Maxime and Darcet, Timoth{\'e}e and Moutakanni, Th{\'e}o and Vo, Huy and Szafraniec, Marc and Khalidov, Vasil and Fernandez, Pierre and Haziza, Daniel and Massa, Francisco and El-Nouby, Alaaeldin and others},
  journal={arXiv preprint arXiv:2304.07193},
  year={2023}
}

@article{ref_retfound,
  title     = {A Foundation Model for Generalizable Disease Detection from Retinal Images},
  author    = {Zhou, Yukun and Chia, Mark A and Wagner, Siegfried K and Ayhan, Maximillian S and Williamson, Dominic J and Struyven, Robbert and Liu, Timing and Xu, Moucheng and Lozano, Mateo G and Woodward-Court, Peter and others},
  journal   = {Nature},
  volume    = {622},
  number    = {7981},
  pages     = {156--163},
  year      = {2023},
  publisher = {Nature Publishing Group}
}

@article{ref_idrid,
  title={Indian diabetic retinopathy image dataset (IDRiD): a database for diabetic retinopathy screening research},
  author={Porwal, Prasanna and Pachade, Samiksha and Kamble, Ravi and Kokare, Manesh and Deshmukh, Girish and Sahasrabuddhe, Vivek and Meriaudeau, Fabrice},
  journal={Data},
  volume={3},
  number={3},
  pages={25},
  year={2018},
  publisher={MDPI}
}

@article{ref_papila,
  title={PAPILA: Dataset with fundus images and clinical data of both eyes of the same patient for glaucoma assessment},
  author={Kovalyk, Oleksandr and Morales-S{\'a}nchez, Juan and Verd{\'u}-Monedero, Rafael and Sell{\'e}s-Navarro, Inmaculada and Palaz{\'o}n-Cabanes, Ana and Sancho-G{\'o}mez, Jos{\'e}-Luis},
  journal={Scientific Data},
  volume={9},
  number={1},
  pages={291},
  year={2022},
  publisher={Nature Publishing Group UK London}
}

@article{ref_messidor2,
  title={Feedback on a publicly distributed image database: the Messidor database},
  author={Decenci{\`e}re, Etienne and Zhang, Xiwei and Cazuguel, Guy and Lay, Bruno and Cochener, B{\'e}atrice and Trone, Caroline and Gain, Philippe and Ord{\'o}{\~n}ez-Varela, John-Richard and Massin, Pascale and Erginay, Ali and others},
  journal={Image Analysis \& Stereology},
  pages={231--234},
  year={2014}
}

@article{ref_refuge,
  title={Refuge challenge: A unified framework for evaluating automated methods for glaucoma assessment from fundus photographs},
  author={Orlando, Jos{\'e} Ignacio and Fu, Huazhu and Breda, Jo{\~a}o Barbosa and Van Keer, Karel and Bathula, Deepti R and Diaz-Pinto, Andr{\'e}s and Fang, Ruogu and Heng, Pheng-Ann and Kim, Jeyoung and Lee, JoonHo and others},
  journal={Medical image analysis},
  volume={59},
  pages={101570},
  year={2020},
  publisher={Elsevier}
}

@article{ref_eyepacs,
  title={EyePACS: an adaptable telemedicine system for diabetic retinopathy screening},
  author={Cuadros, Jorge and Bresnick, George},
  journal={Journal of diabetes science and technology},
  volume={3},
  number={3},
  pages={509--516},
  year={2009},
  publisher={SAGE Publications}
}

@article{ref_bilateral1,
  title={Automated diabetic retinopathy detection based on binocular siamese-like convolutional neural network},
  author={Zeng, Xianglong and Chen, Haiquan and Luo, Yuan and Ye, Wenbin},
  journal={IEEE access},
  volume={7},
  pages={30744--30753},
  year={2019},
  publisher={IEEE}
}

@inproceedings{ref_bilateral2,
  title={Two eyes are better than one: Exploiting binocular correlation for diabetic retinopathy severity grading},
  author={Qian, Peisheng and Zhao, Ziyuan and Chen, Cong and Zeng, Zeng and Li, Xiaoli},
  booktitle={2021 43rd Annual International Conference of the IEEE Engineering in Medicine \& Biology Society (EMBC)},
  pages={2115--2118},
  year={2021},
  organization={IEEE}
}

@article{ref_bilateral3,
  title={Siamese network based fine grained classification for Diabetic Retinopathy grading},
  author={Nirthika, Rajendran and Manivannan, Siyamalan and Ramanan, Amirthalingam},
  journal={Biomedical Signal Processing and Control},
  volume={78},
  pages={103874},
  year={2022},
  publisher={Elsevier}
}

@article{ref_oct1,
  title={Analysis of the asymmetry between both eyes in early diagnosis of glaucoma combining features extracted from retinal images and octs into classification models},
  author={Rodriguez-Robles, Francisco and Verdu-Monedero, Rafael and Berenguer-Vidal, Rafael and Morales-Sanchez, Juan and Selles-Navarro, Inmaculada},
  journal={Sensors},
  volume={23},
  number={10},
  pages={4737},
  year={2023},
  publisher={MDPI}
}

@misc{ref_odir,
  title        = {Ocular Disease Intelligent Recognition ({ODIR-2019})},
  author       = {{Peking University} and {Shanggong Medical Technology}},
  year         = {2019},
  howpublished = {\url{https://odir2019.grand-challenge.org/}}
}

@misc{ref_aptos2019,
  author       = {Karthik and Maggie and Sohier Dane},
  title        = {APTOS 2019 Blindness Detection},
  year         = {2019},
  howpublished = {\url{https://kaggle.com/competitions/aptos2019-blindness-detection}},
  note         = {Kaggle}
}

@misc{ref_who_vision,
  author       = {{World Health Organization}},
  title        = {World Report on Vision},
  year         = {2019},
  howpublished = {\url{https://www.who.int/publications/i/item/9789241516570}}
}

@misc{ref_retina,
  author       = {{Jr2ngb}},
  title        = {Retinal Cataract Dataset},
  year         = {2020},
  howpublished = {\url{https://www.kaggle.com/datasets/jr2ngb/cataractdataset}}
}

@article{ref_dinov3,
  title={Dinov3},
  author={Sim{\'e}oni, Oriane and Vo, Huy V and Seitzer, Maximilian and Baldassarre, Federico and Oquab, Maxime and Jose, Cijo and Khalidov, Vasil and Szafraniec, Marc and Yi, Seungeun and Ramamonjisoa, Micha{\"e}l and others},
  journal={arXiv preprint arXiv:2508.10104},
  year={2025}
}

@article{ref_medmamba,
  title={Medmamba: Vision mamba for medical image classification},
  author={Yue, Yubiao and Li, Zhenzhang},
  journal={arXiv preprint arXiv:2403.03849},
  year={2024}
}

\end{document}